\documentclass[sigconf,screen,leftjustify]{acmart}
\renewcommand\footnotetextcopyrightpermission[1]{}
\setcopyright{rightsretained}%{acmcopyright}
\copyrightyear{2021}
\acmYear{2021}
\acmDOI{}%10.1145/1122445.1122456}

\usepackage{fancyhdr}
\acmConference[KDD '21: MIS2 Workshop]{Knowledge Data Discovery: Misinformation and Misbehavior Mining on the Web (MIS2) Workshop}{August 15, 2021}{Virtual}
\acmBooktitle{The 27th ACM SIGKDD Conference on Knowledge Discovery and Data Mining (KDD'21): Misinformation and Misbehavior Mining on the Web (MIS2) Workshop, August 15, 2021, Virtual}
\usepackage{times}
\usepackage{epsfig}
\usepackage{graphicx}
\usepackage{amsmath}

\usepackage{amssymb}
\usepackage{tabularx}
\usepackage{tablefootnote}
\usepackage{subfigure}
\usepackage{booktabs}
\usepackage{extpfeil}

\usepackage{algorithm,algorithmic}
\usepackage{newtxmath} % or: \usepackage{newpxmath}
\usepackage{ulem}

\def\Reals{\mathbb{R}}

\def\vec#1{{\bf#1}}
\def\mat#1{{\bf#1}}
\def\ten#1{{\boldsymbol{\mathcal#1}}}

\def\tp{^\mathrm{T}}

\def\assign{\mathrel{\mathop:}=}
\def\approx{\simeq}

\def\mode#1{_{\mbox{\tiny\rm#1}}}

\def\matize#1#2{{\mat#1}\mode{$[#2]$}}

\def\pinv#1{^{\dag\lower2pt\hbox{\hskip-1pt\hbox{\tiny${}#1$}}}}
\begin{document}
\title{Deepfake Representation with Multilinear Regression}

{\author{Sara Abdali%\vspace{+.175in}
}% \thankssymb{*}  }
%\thanks{\thankssymb{*} The authors contributed equally to this paper. }
\email{sabda005@ucr.edu}
\affiliation{%\vspace{+.15in}
  \institution{
  University of California, Riverside}}
 %\protect\label{X}
 }
\author{M. Alex O. Vasilescu }%\thankssymb{*}}
 
\email{maov@cs.ucla.edu}

\affiliation{%
 \institution{University of California, Los Angeles}
 \institution{Tensor Vision, Los Angeles}
 }
\author{Evangelos E. Papalexakis%\vspace{+.175in}
} \email{epapalex@cs.ucr.edu}
\affiliation{
  \institution{
  University of California, Riverside}
  }

\begin{abstract}
Generative neural network architectures such as GANs, may be used to generate synthetic instances to compensate for the lack of real data. However, they may be employed to create media that may cause social, political or economical upheaval. One emerging media is "Deepfake". Techniques that can discriminate between such media is indispensable. In this paper, we propose a modified multilinear (tensor) method, a combination of linear and multilinear regressions for representing fake and real data. We test  our approach by representing Deepfakes with our modified multilinear (tensor) approach  and perform SVM classification with encouraging results.

\end{abstract}
\maketitle

\section{Introduction}
Recent advances in Generative Adversarial Networks (GANs) and Convolutional Neural Networks (CNNs) embedded in applications like Zao\footnote{\tiny https://www.zaoapp.net/}, DeepFakes web $\beta$\footnote{ \tiny https://deepfakesweb.com/}, Face Swap by Microsoft\footnote{\tiny https://www.microsoft.com/en-us/garage/profiles/face-swap/}, DeepFaceLab\footnote{\tiny https://awesomeopensource.com/project/iperov/DeepFaceLab} etc. have led to a broad usage of AI-synthesized media a.k.a. "Deepfake"~\footnote{The term Deepfakes has been widely used for deep learning generated media, but it is also the name of a specific manipulation technique in which face of one person is replaced by another one. To distinguish these, we denote said method by DeepFakes in the entire paper.}. Other automated manipulation techniques are Face2Face, FaceSwap, NeuralTextures, and FaceShifter~\cite{roessler2019faceforensicspp}.
\par Due to the potential misuse of Deepfakes e.g., fake pornography, fake news, and financial or political fraud, they have become a major public concern. Thus, different techniques have been introduced to discriminate Deepfakes from pristine videos.

%Interested reader is refer to~\cite{roessler2019faceforensicspp} for more details on the aforementioned methods.

\par Prior Deepfakes detection can be categorized as ~\cite{2020_SurveyDeepFake} approaches that classify based on (a) physical or physiological causal factors which are not well presented in Deepfakes e.g., eye blinking~\cite{eyeblinking} and 
heart rate\cite{heart_rate}, or (b) artifacts in imaging factors e.g., relative head pose to the camera position\cite{HeadPose}, and (c) data-driven techniques that do not leverage specific cues and directly train a deep learning model on a large set of real and Deepfake videos~\cite{MesoNet,Xception}. 

\par From the first category, we can mention ~\cite{HeadPose}, where Yeng et al. propose using the inconsistencies in head poses to detect the Deepfakes. More precisely, 3D head poses cue is leveraged to estimate errors introduced by splicing process which synthesizes source face region into the target one. The eye blinking cue is anther physiological signal which is not well presented in Deepfakes and Li et al. take advantage of it for discriminating the Deepfakes~\cite{eyeblinking}. More recently, a novel cue has been introduced that considers the heart rate measured by remote photoplethysmography (rPPG) to analyze color changes in the human skin, which is a signal for the presence of blood under the tissues ~\cite{heart_rate}.
\begin{figure}[!bt]
\vspace{+.075in}
    \begin{center}
  \includegraphics[width = .9\linewidth]{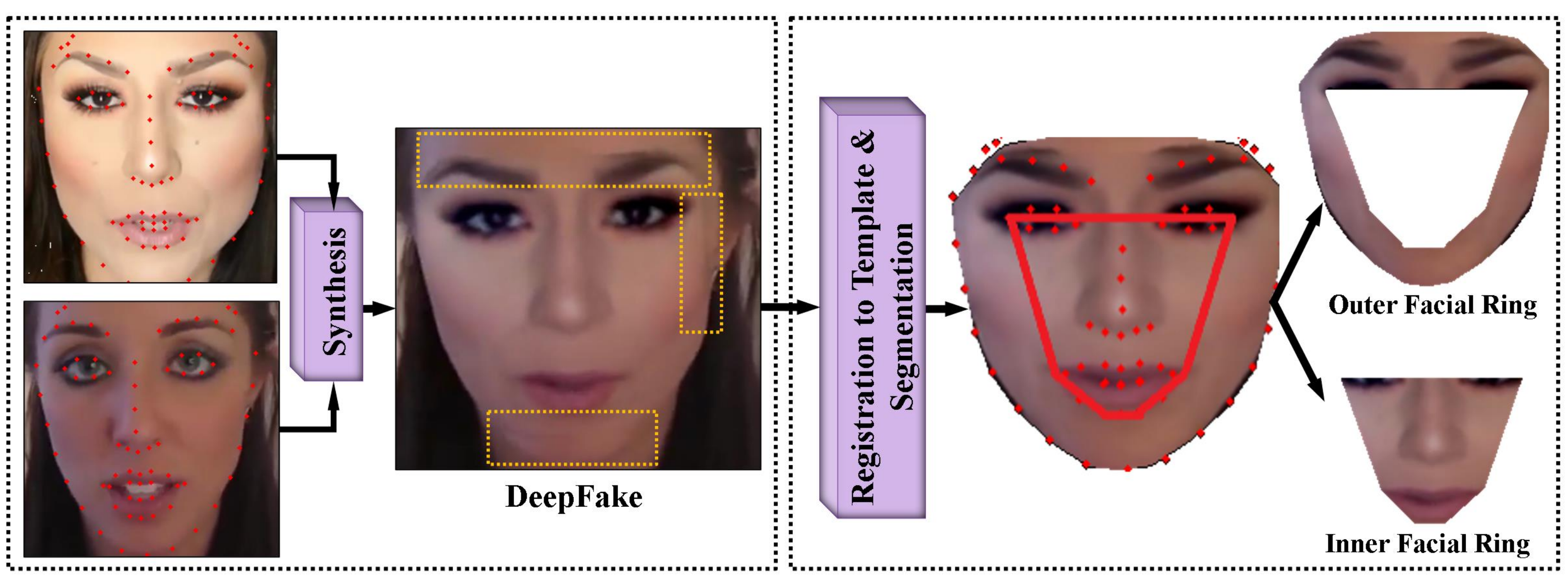}
  \vspace{-.1in}
    \end{center}
    \caption{    Deepfake technique replaces a person's appearance in an existing image or video with someone else's appearance~\cite{roessler2019faceforensicspp}. This process introduces artifacts specially around the cropping boundaries estimated by facial landmarks. We propose segmenting the output face into inner and outer facial rings. The artifacts are mainly concentrated in outer facial ring.
    \iffalse
    Deepfake technique replaces a person's appearance in an existing image or video with someone else's appearance. %Each image is the composite consequence of different causal factors.  
    Unmitigated differences in 
    scene structure, illumination, or viewing conditions of the source images results in image artifacts. %We propose to analyze the segmented the face parts, the inner and outer facial rings. 
    We hypothesize that the artifacts are located primarily in the outer face ring.
    \fi
    }
    \label{fig:process}
    \vspace{-.2in}
\end{figure}
\par As an example of the second category, we can refer to the work in ~\cite{artifact} where the distinctive feature is the introduced face warping artifacts. In this work, Li et al. discuss limitation of early Deepfake generators which produce images of limited resolutions and transformation of this images leaves certain distinctive artifacts in the Deepfake videos. In addition, in ~\cite{Saturation_Cues}, McCloske et al. analyze the structure of the generator network of a GAN and show how the network’s treatment of exposure is markedly different from a real camera. They propose leveraging frequency of over-exposed pixels as a feature for this cue to discriminate GAN-generated media from camera imagery.
\par However, the vast majority of proposed methods for Deepfake detection fall into the third category, i.e., data-driven approaches. For instance, in ~\cite{HybridLSTM_Amit} a hybrid Long Short Term Memory Network (LSTM) and Encoder-Decoder architecture is introduced to detect forgeries in images. In another work ~\cite{Xception}, a novel CNN network inspired by inception is introduced, where inception modules have been replaced with depth wise separable convolutions. Another example of this category is the work proposed in ~\cite{MesoNet}, where two networks are presented, both with a low number of layers to focus on the mesoscopic properties of the images. ~\cite{RNN+CNN}, ~\cite{alignment} and~\cite{Multi_task_Learning} are other instances of data-driven approaches which leverage (Recurrent Neural Networks (RNNs), capsule networks and CNN networks for detection of Deepfakes. Lastly, there are works that take advantage of CNNs and RNNs simultaneously to capture both frame level and sequence level information~\cite{RNN+CNN,HybridLSTM_Amit,alignment}.

 \par The first two categories, which mainly leverage feature extraction and image pre-processing techniques to some extent provide interpretability for the classification result which is a key factor for explainable and trustworthy AI. For instance, the predictive model is built upon differences in the nature of pixels~\cite{Saturation_Cues} or an estimation of regions with high concentration of artifacts~\cite{artifact}. However, the black box methods of the third category, while being highly accurate, do not provide any interpretation for the classification output. 
  Thus, it is not clear if the video is classified as Deepfake due to the difference in the frame by frame movement or because of spatial artifacts or both. Moreover, there is no information about regions of interest and causes of the artifacts e.g., warping artifacts or artifacts in head adjustment, that discriminate Deepfakes from real videos.

\par %To do so, w
%We focus on %the study of discrepancies generated by DeepFakes technique. 
We hypothesize that DeepFakes contain artifacts localized either in transition areas between facial images, or contain discrepancies in the overall facial appearance.  We concentrate our analysis on the transition areas of the face henceforth referred to as outer facial ring, Figure~\ref{fig:process}.
%We propose canceling inner parts and 
We segment the outer ring from a facial image that has been registered to a template based on facial landmarks detected by a pretrained model~\cite{landmark}. The outer ring is analyzed with a modified face recognition tensor model~\cite{Vasilescu02,Vasilescu05} that computes real and fake data representations.

\par %To this end, w
We employ a multilinear a.k.a. tensor framework  which decomposes basis components of outer facial rings into real and fake class representations. Later on we leverage the derived representation of classes to classify the test frames using a linear SVM. Summarily, our major contributions are as follows: 
\begin{itemize}
\item \textbf{Segmenting face into regions of interest}: We propose Segmenting face into facial parts and leverage parts with high concentration of artifacts to distinguish Deepfakes.
    \item \textbf{Proposing a multilinear representation of Deepfakes for classification:} we employ a multilinear approach to represent Deepfake and real class information and then leverage them for classification. 

\end{itemize}
\section{Background}
In this section, we discuss the relevant tensor algebra~~\cite{Delathauwer00a},~~\cite{Delathauwer00b},~~\cite{Vasilescu02,Vasilescu05},~\cite{KoBa09},~\cite{Papalexakis:2016},~\cite{Tensor}. We will follow the notation of Table~\ref{notation}

\subsection{Multilinear (tensor) framework}
A data tensor $\ten{D} \in  {\rm I\!R}^{I_1\times I_2\times\dots\times I_M} $ is a multi-way array. In fact, when an array has three or more than three dimensions, we call it a tensor. The dimensions of a tensor are usually referred to as modes.
\subsection{Singular Value Decomposition (SVD) and\\Principle Components Analysis (PCA)}
In linear algebra, we factorize a matrix $\mat{D} \in {\rm I\!R}^{I_1\times I_2}$ using Singular Value Decomposition (SVD) as follows:
\begin{equation}
\mat{D} =\mat{U}\mat{\Sigma} \mat{V}\tp
\label{SVD}
\end{equation}
where the columns of $\mat U \in {\rm I\!R}^{I_1\times r}$ and $\mat V \in {\rm I\!R}^{I_2\times r}$ are orthonormal and
$\mat{\Sigma \in {\rm I\!R}^{r\times r}}$ is a diagonal matrix with positive real entries know as singular values. The rank $R$, SVD decomposition represents a matrix as following equation: %~\cite{KoBa09,Papalexakis:2016,Tensor}:

\begin{equation}
\mat{D}
\approx \sum_{r=1}^{R} \sigma_r  \vec{u}_r \circ \vec{v}_r 
\end{equation}
\par Rewriting equation \ref{SVD} in conventional linear algebra, the Principal Components Analysis (PCA) is: 
\begin{equation}
\mat{D} =\underbrace{\mat{U} 
}_\text{Basis} \underbrace{\mat{\Sigma}\mat{V}\tp}_{\text{Coefficient}}
\end{equation}
\begin{table}[t!]
\centering
\begin{tabular}{
p{1.5cm}p{5.5cm}}
 \multicolumn{2}{l}{} \\
 \hline
 \textbf{\small Symbol} &\textbf{\small Definition}\\
 \toprule
 \textbf{$\ten{D}$}, $\mat{D}$, $\vec d$& Tensor, Matrix, vector\\
 $\ten{D}\pinv{m}$& Mode-m tensor pseudo-inverse of $\ten{D}$ \\
 $\matize{D}{m}$& Mode-$m$ tensor matrixizing \\
 $\times\mode m$&Mode-$m$ product\\
 $\circ$&Outer product\\
\bottomrule
\end{tabular}
\caption{Symbols and Definition}
\label{notation}
\vspace{-.2in}
\end{table}
\subsection{ Mode-$M$ Matrixizing a Tensor}
 TMode-$m$ matrixizing of tensor $\ten{D} \in  {\rm I\!R}^{I_1\times I_2\times\dots\times I_M} $ is defined as the matrix $\matize{D}{m} \in {\rm I\!R}^{I_m\times(I_1\dots I_{m-1}I_{m+1}\dots I_M)} $
 where the parenthetical ordering
indicates that column vectors are ordered by sweeping
indices of all other modes through their ranges~\cite{Vasilescu09}. Therefore:
\begin{equation}
 \begin{aligned}
   [\mat{D}]_{jk}=&a_{i_1\dots i_m\dots i_M}
   \text{\hspace{5pt}where} \\
   &j=i_m \text{\hspace{5pt}and\hspace{5pt}} k=1+\sum_{n=0 ,n\neq m}^{M} (i_n-1)\prod_{l=0 ,l\neq m}^{n-1} I_l
    \end{aligned}
\end{equation}
\par
\begin{figure}[tb]
    \begin{center}
  \includegraphics[width = 1\linewidth]{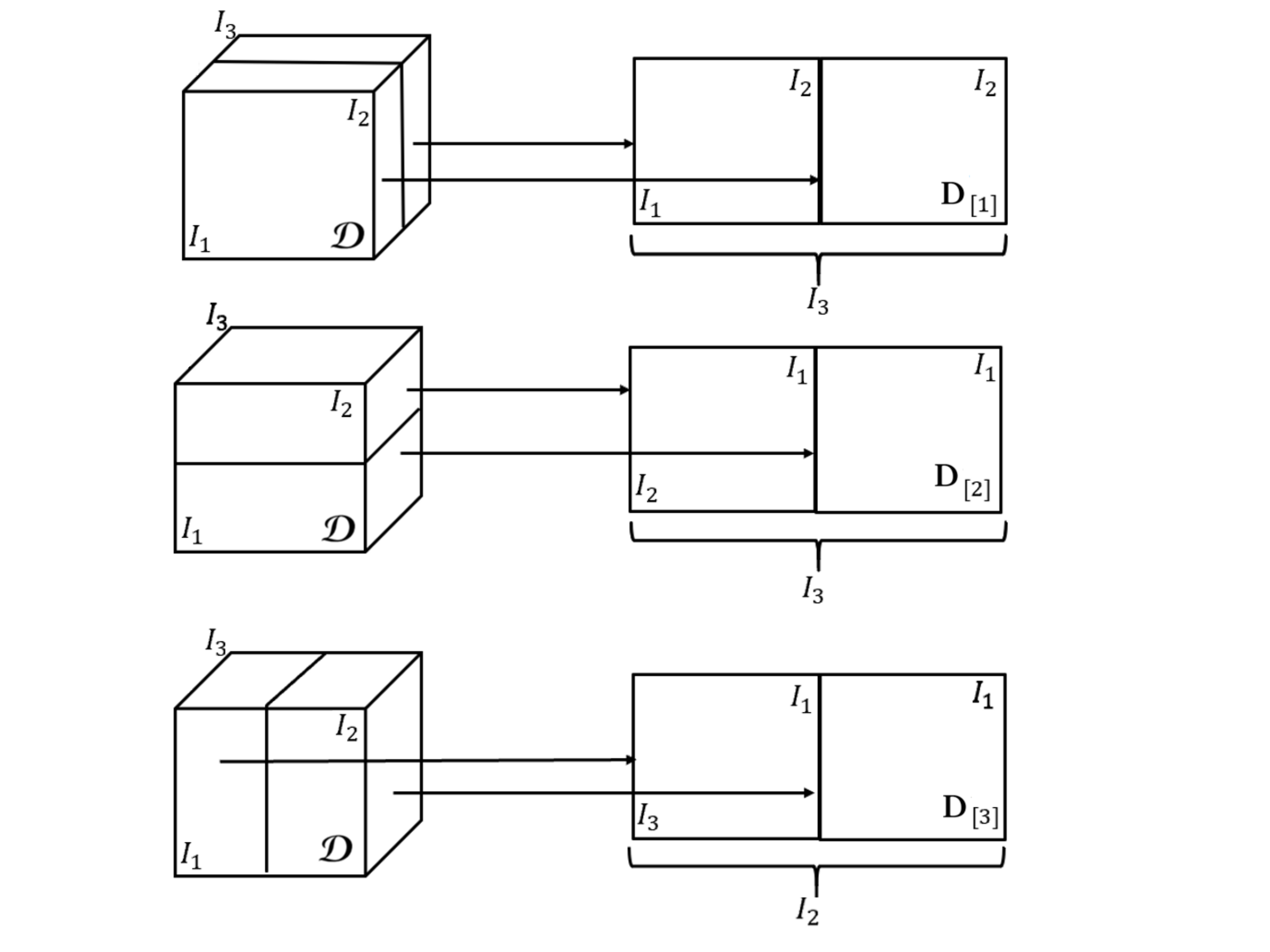}
  \vspace{-.4in}
    \end{center}
    \caption{Matrixizing a 3-mode tensor}
    \label{fig:unfolding}
      \vspace{-.25in}
\end{figure}

 A $3$-mode tensor may be metricized in three different ways by stacking first, second and third mode slices which are illustrated in Figure.\ref{fig:unfolding}.

\subsection{Mode-$M$ product of a matrix and a tensor}
The mode-$m$ product~\cite{Vasilescu09,Carroll70,Delathauwer00a} of a tensor
$\ten{D} \in  {\rm I\!R}^{I_1\times I_2\times\dots I_m\times\dots\times I_M} $ and matrix $\mat{A} \in  {\rm I\!R}^{J_m\times I_m}$ denoted by $\ten{D}\times\mode m\mat{A}$ is a tensor of size ${\rm I\!R}^{I_1\times I_2\times\dots J_m\times\dots\times I_M}$ where the entries are calculated as
\begin{equation}
\begin{aligned}
   [\ten{D}\times\mode m\mat{A}]_{i_1\dots i_{m-1}j_m i_{m+1}\dots i_M}
   = \sum_{i_m}^{} d_{i_1 i_2 \dots i_{m-1} i_m i_{m+1}\dots i_M}a_{j_mi_m} 
   \end{aligned}
\end{equation}

The mode-$M$ product is interchangeably denoted by matrix multiplication and tensor multiplication as follows: 

\begin{equation}
   \ten{B}=\ten{D}\times\mode n\mat{A}  	\xtofrom[\text{tensorizing}]{\text{matrixizing}}
   \matize{B}{m}=\mat{A}\matize{D}{m}
\end{equation}
\subsection{$M$-mode SVD}
We can define SVD decomposition in terms of n-mode product as follows: %~\cite{KoBa09,Papalexakis:2016,Tensor}:
\begin{equation}
\mat{D} =\mat{\Sigma}\times\mode 1\mat{U} \times\mode 2\mat{V}
\end{equation}
In multilinear algebra there is a generalization of SVD know as multilinear SVD~\cite{Delathauwer00a,Delathauwer00b} or $M$-mode SVD~\cite{Vasilescu02,Vasilescu05} which decomposes an $M$-mode tensor $\ten{D}$ into the $M$-mode product of orthonormal spaces: %~\cite{KoBa09,Papalexakis:2016,Tensor,DeLathauwer00,Vasilescu02,Vasilescu05}:
\begin{equation}
    \ten{D} \approx \ten{Z}\times\mode 1 \mat{U}\mode{1} \times\mode 2\mat{U}\mode{2}\dots\times\mode M\mat{U}\mode{M}
    \label{eq:hosvd}
\end{equation}
where 
$\ten{Z}$ is the core tensor that governs the interaction between the orthonormal mode matrices, $\mat U\mode m$.  The core tensor is analogues to the singular value matrix $\Sigma$ but unlike the $\Sigma$ the core tensor is not always diagonal~\cite{KoBa09,Papalexakis:2016,Tensor}. 
\par
\begin{figure}[!t]
    \begin{center}
  \includegraphics[width = .95\linewidth]{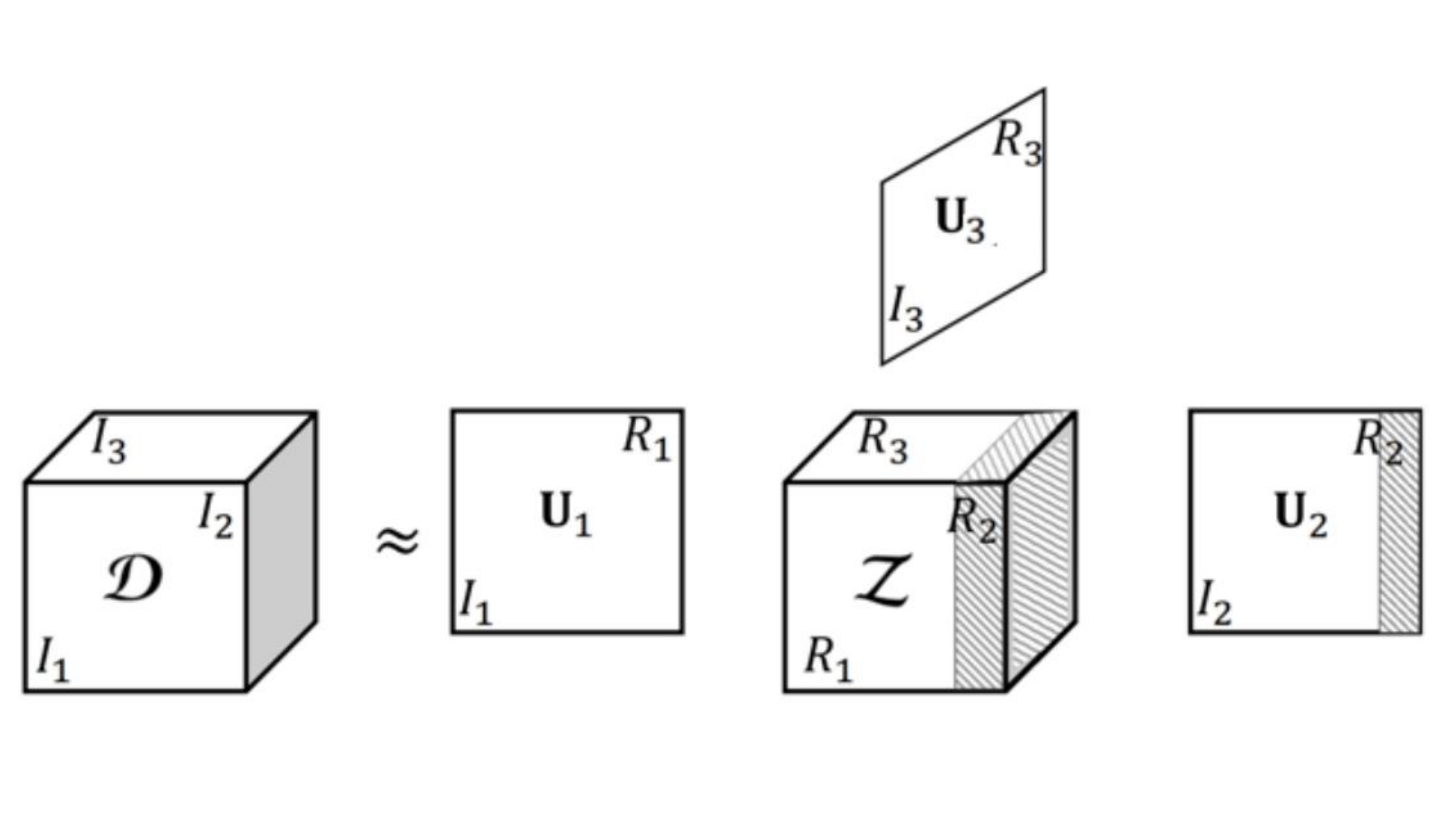}
 % \vspace{-.5in}
    \end{center}
    \caption{$M$-mode SVD decomposition of a 3-mode tensor. Some of the singular values corresponding to the components of the second factor matrix i.e., second mode of tensor $\ten{D}$ are truncated.}
    \label{fig:truncation}
      \vspace{-.2in}
\end{figure}

 The $M$-mode SVD of a 3-mode tensor is demonstrated in Figure \ref{fig:truncation}.
$\mat{U}_i$ is approximated by left singular vectors of truncated SVD decomposition of $\matize{D}{i}$.
Meanwhile, since $\mat{U}_i$ is orthonoramal, we have $\mat{U}_m^{-1}=\mat{U}_m^T$ and the core tensor $\ten{Z}$ is estimated as follows:
\begin{eqnarray}
    \ten{Z} &=& \ten{D}\times\mode 1 \mat{U}\mode{1}^{-1}\times\mode 2\mat{U}\mode{2}^{-1}\dots\times\mode M\mat{U}\mode{M}^{-1}\\
&=& \ten{D}\times\mode 1  \mat{U}\mode{1}^{T}\times\mode 2\mat{U}\mode{2}^{T}\dots\times\mode M\mat{U}\mode{M}^{T}
\end{eqnarray}

\section{Proposed Method}
A DeepFake is a synthesizing product of two real faces. More precisely, in DeepFake generation process, face of a real person a.k.a. target is synthesized by another face a.k.a., source. This process, usually introduces some artifacts, specially around the cropping edges of source face including eyes and eyebrows Figure \ref{fig:process}. Due to the fact that a Deepfake face is a mixture of source and target faces, Sometimes it is not distinguishable from the source and this similarity results in misclassification of the video. In this work, we propose to segment faces into parts henceforth referred to as facial inner and outer rings Figure \ref{fig:process}. We define the outer ring as a facial part that comprises the blending boundaries that are mostly the non-facial pixels. We leverage this remaining region i.e., outer ring which has the highest concentration of introduced artifacts as a cue for Deepfakes detection. An example of this process is demonstrated in Figure. \ref{fig:process}. This cue is very promising specially when the manipulation masks are not available.
\par In what follows, we discuss our proposed multilinear pipeline for detecting the Deepfakes. 
 
\subsection{Step 1: Vectorizing video frames}
Vasilescu~\cite[Appndix A]{Vasilescu09} argues that in most cases,
it is preferable to vectorize an image and treat it as a single observation rather
than a collection of independent column/row observations. %and how it helps improving the performance of face recognition. 
By vectorizing an image,
we treat an image as a point in high dimensional pixel space and calculate all possible combinations of pixel statistics, both near and faraway statistics. 
On the other hand, when we consider an image as a matrix, every image column (row) is treated as an independent observation, and column (row) covariances are computed.
%each pixel only covaries with pixels of the same row and the same column. %and we are not able to capture all possible pairwise statistics ~\cite{Vasilescu09}. %,MPCAandMICA}. 
Having this in mind, we also follow the same strategy and vectorize the frames and create a vector for each one of the video frames in the dataset.

\subsection{Step 2: Finding eigenfaces of each class}
Eigenfaces are eigenvectors when the images are human face. The eigenfaces are derived from the covariance matrix of the pixel distribution over the high dimensional face space.  The eigenfaces represent a basis set of all faces used to construct the covariance matrix. So far, the eigenfaces have been successfully leveraged for many facial image related tasks~~\cite{Turk91b}. Leveraging eigenfaces allows for dimensionality reduction such that a smaller set of basis vectors represent the original training faces. Classification could be achieved by comparing how different faces are represented by the basis set of the corresponding class.

Based on principle component terminology, the eigenfaces are equal to basis vectors of PCA decomposition. Therefore, by staking the vectorized frames of each class, we create two separate matrices and decompose them using SVD to capture the eigenfaces of the corresponding class as follows:
\begin{eqnarray}
\mat{D}\mode {real}&=&\mat{U}\mode {real}\mat{\Sigma}\mode {real}\mat{V}\mode {real}\tp=\mat{B}\mode {real}\mat{V}\mode {real}\tp\\
%\end{equation}
%\begin{equation}
\mat{D}\mode {fake}&=&\mat{U}\mode {fake}\mat{\Sigma}\mode {fake}\mat{V}\mode {fake}\tp=\mat{B}\mode {fake}\mat{V}\mode {fake}\tp
\end{eqnarray}
Where $\mat{B}\mode {real}$ and $\mat{B}\mode {fake}$ are basis matrices and $\mat{V}\mode {real}$ and $\mat{V}\mode {fake}$ are the normalized coefficient matrices of the corresponding classes. 

\subsection{Step 3: Leveraging tensor framework to decompose eigenfaces into underlying factors}
As seen in previously mentioned tensor is an effective framework for decomposing a set of observation into underlying factors. After reducing the dimentionality of observations using eigenface representation of the classes, we propose leveraging a three-mode tensor where the first mode i.e.,  measurement mode represents the pixels of an eigenface, the second mode corresponds to the eigenfaces and the third mode is the class mode i.e., DeepFake vs. real.
We propose using an $M$-mode SVD which as we discussed earlier decomposes a tensor into $M$ orthonormal matrices ($M=3$), and a core tensor which governs the interaction between these spaces. 
Since the first mode is the measurement mode, We only calculate the $M$-mode SVD of the tensor by flattening the second and the third modes as follows:%~~\cite{MPCAandMICA,Multilinear_Projection2007,Vasilescu09}:
\begin{eqnarray}
    \ten{D} &\approx& \ten{Z}\times\mode 1 \mat{U}\mode {p} \times\mode 2\mat{U}\mode {f}\times\mode 3\mat{U}\mode {c}\\
&=& \ten{T}\times\mode 2\mat{U}\mode {f}\times\mode 3\mat{U}\mode{c}
\end{eqnarray}
where the $\mat{U}\mode{c}$ %is the third factor matrix corresponding to the class representations and 
comprises underlying vector representation of original and fake classes.% This tensor is depicted in Figure  \ref{fig:tensor}. 
 Moreover, the core tensor $\ten{T}$ is the signature of this dataset and shows interactions of orthonormal subspaces. Later on, we leverage this signature to project the test frames into the subspaces we derive here.
 \subsection{Step 4: Embedding the class representations in a higher three dimensional space}
Applying $M$-mode SVD results in a mode matrix $\mat{U}\mode{c} \in {\rm I\!R}^{2\times 2}$ that spans the class representations. %which means we have a two dimensional vector representation for each class. %However, such extreme dimensionality reduction might not be representative enough to  discriminate all categories of real and fake frames. 
We embed the vector class representations into a higher dimensional space to increase the class separability of the test data. %such that extended dimensions reflect the nature of classes.
%Therefore, to increase the linear separability of our class representations, w
We embed the row vectors of $\mat{U}\mode{c}$ into $\Reals^3$,
%into into $\mat{U}\mode{c}^+ \in\Reals^{2\times3}$ i.e., a higher three dimensional space 
setting the third coordinate of the real and fake class to $+1$ and $-1$ respectively, and normalizing the vector length to $1$. %Later on, we will show how this process makes the class representations more separable.
\subsection{Step 5: Multilinear projection of an incoming frame into the orthonormal vector spaces}
As mentioned above, the core tensor of each decomposition is the signature of the  decomposed space which governs the interaction of constituent factors.
we leverage the core tensor and perform a  multilinear projection of the incoming frame into the subspaces we derived in the previous step. Let say we have the vectorized frame $\vec d$. If $\vec d$ is supposed to be in the same subspaces we derived, then
\begin{equation}
    \Vec{d}=\ten{T}\times\mode 2 \Vec{f}\tp\times\mode 3 \Vec{c}\tp 
    \label{eq:observation_model}
\end{equation}
where the vectors $\Vec{f}$ and $\Vec{c}$ are the coefficient vector representations of a video frame $\vec d$ in the orthonormal subspaces that are governed by the extended core tensor $\ten{T}$. The goal is to find out weather the class coefficient vector $\Vec{c}$ is more similar to the vector representation of real class or Deepfake class. To this end, we estimate $\Vec{c}$ representation vector by employing the multilinear projection algorithm~\cite{Vasilescu11,Multilinear_Projection2007}
that decomposes a vectorized observation, $\vec d$ into a set of latent vector representation, $\vec r\mode n$ that corresponds to the constituent factors of data formation.  The basic multilinear projection is the $M$-mode SVD/CP decomposition of $\ten{T}\pinv1\times\mode 1\vec d\tp$ which can be expressed mathematically as 

\begin{eqnarray}
%\mbox{Multilinear Projection:}\hspace{+1.5in}&&\\
%\vec{r}\mode {s} \circ\vec{r}\mode {c}\Leftarrow 
\underbrace{\mbox{\small $M$-mode SVD/CP} \left( \ten{T}\pinv 1\times\mode 1\vec d\tp\right)}_{\mbox{Multilinear Projection}} \approx  \vec{r}\mode{f} \circ \vec{r}\mode{c}%\hspace{+.05in}
\hspace{+.05in}
\Rightarrow
\hspace{+.05in}
%\ten{T}\pinv1\times\mode 1\vec d\tp &=&\ten R\nonumber\\
%\ten R\hspace{+1.25in} &\approx&\vec{r}\mode {var} \circ \vec{r}\mode {cls}\nonumber
%\ten{R} %&\Big\Downarrow& \mbox{\smaller $M$-mode SVD $(\ten R)$}\nonumber\\
\vec d\approx %\underbrace{
\left(\ten T\times\mode 2 \vec{r}\mode{f}\tp \times\mode 3 \vec{r}\mode{c}\tp\right)\nonumber
%}_{\text{{ $M$-mode SVD / CP }} \left( \ten R \right)}\nonumber
%&&\hspace{+.4in}=\underbrace{\left(\vec{r}\mode {s} \circ\vec{r}\mode {c}\right)}_{\mbox{{\smaller CP}} \left( \ten R \right)}
\end{eqnarray}
where %$\ten{P}=
$\ten T\pinv{1}$ is mode-1 pseudo-inverse of $\ten T$ that in matrix notation is expressed as $\matize{T}{1}\pinv{}$, 
and $\vec r\mode{c}, \vec r\mode{f}$ 
are estimates of vectors $\vec{c}$ and $\vec{f}$ from eq.(\ref{eq:observation_model}), respectively.

%\vspace{-.2in}
\subsection{Step 6: Classifying an incoming frame}
Up to this step, we have the vector representation of each classes in addition to class coefficients of the incoming frame. We use a linear Support Vector Machine (SVM) and estimate the decision boundaries using validation frames and then leverage the defined boundaries for classification of test frames. An overview of the proposed approach is demonstrated in Algorithm \ref{alg:deepfake}. 

\subsection{Dimensionality reduction in step 3}
As mentioned earlier, factor matrix $\mat{U}_f$ comprises underlying structures of basis vector continent.
Despite the fact that we construct our predictive model by approximating discriminating regions, still there are many shared components which getting rid of them make the model more distinguishable. Since we are interested in noisy regions i.e., artifacts, we propose truncating components of the core tensor $\ten{T}$ which correspond to top values of $\mat{U}_f$ and keeping lower value components as representatives of noisy parts.
 In the next section, we will show how this truncation boosts the classification performance of the proposed framework. An example of truncating components corresponding to the second mode of a 3-mode tensor is depicted in Figure. \ref{fig:truncation}.  
\begin{algorithm}[tb!]
%\SetAlgoLined
%\small
%\LinesNumbered
%\algorithmicrequire{A set of DeepFake/original videos \& an unlabeled video $D\mode{real},\mat D\mode{fake]$}\\
%\algorithmicensure{Label of video $d$}\\
%\midrule
%$\backslash \backslash$ \textbf{Step 1}\\
%Vectorize all frames, original and manipulated\\
%$\backslash \backslash$\textbf{Step 2}\\
$\mat{Input:}$ $\mat D\mode{real}, \mat D\mode{fake}$ were centered by subtracting the mean of the real training data,
\begin{enumerate}
\item 
    Preprocessing and data tensor organization:\\
    $[\mat U\mode{real},\mat S\mode{real},\mat V\mode{real}] \Leftarrow \mbox{svd}(\mat D\mode{real})$\\
    $[\mat U\mode{fake},\mat S\mode{fake},\mat V\mode{fake}] \Leftarrow \mbox{svd}(\mat D\mode{fake})$\\
    $\ten{D}(:,:,1)=\left[\mat{U}\mode{real}\mat{S}\mode{real}\right]$\\
    $\ten{D}(:,:,2)=\left[\mat{U}\mode{fake}\mat{S}\mode{fake}\right]$
\item 
Training data decomposition:\\
%$\matize{D} x=[\mat{B}\mode {orig},\mat{B}\mode {fake}]$\\
$\ten{T} \times\mode{2} \mat U\mode{f}\times\mode{3}\mat{U}\mode {c}\Leftarrow\text{$M$-mode SVD}(\ten{D})$
\item
Embed the class representations in the higher three dimensional space and set the third coordinate of the real and fake class to $+1$ and $-1$ respectively. Hence, $\mat U\mode c \in\Reals^{2\times2}$ now has dimensionality $\Reals^{2\times 3}$. Normalize the rows of $\mat U\mode c$ to have length $1$.
\item Computer the extended core
\begin{equation}
\ten T\assign\ten D\times\mode 2 \mat U\mode f\tp \times\mode 3 \mat U\mode c\pinv{}
\end{equation}
\item Centering: validation and test data is centered by subtracting the mean of the real training data.
\item
{Test data decomposition of a centered $\vec d\mode{test}$ :}
\begin{equation}
\vec d\mode{test}\approx\ten T \times\mode 2 \vec r\mode{f}\tp \times\mode 3 \vec r\mode{c}\tp\Leftarrow \mbox{Multilinear Projection}(\ten T, \vec d\mode{test})\nonumber
\end{equation}
%$\mat{T}\mode {[1]}=\text{Unfold}(\ten{T},1)$\\
%$\text{pseudoInverse}(\mat{T}\mode {[1]}^{+})$\\
%$\mat{T}\mode {[1]}^{+}=\mat{T}\mode {[1]}$\\
%$\ten{P}=\text{Fold}(\mat{T}\mode {[1]}^{+},1)$\\
%$\ten{R}\approx \ten{P}\times\mode 1 d\tp$  \\
%$[\ten{Z} ,\mat{R}\mode {s},\mat{R}\mode {c}]=\text{$M$-mode SVD}(\ten{R})$\;\\
\item  Finding linear SVM decision boundaries using validation set
\item classifying all $\vec d\mode{test} \in$ test set
%Nearest Neighbor Classification:\\

%Computing a cosine similarity measure between $\vec r\mode{c}$ and each of the two rows of $\mat U\mode{c}$, \ie $\vec c\mode{i=real}\tp$ and $\vec c\mode{i=fake}\tp$
%\begin{equation}\argmax_{i} \frac{\vec c\mode{i}\tp \vec r\mode{c}}{\|\vec c_i\| \|\vec r_{c}\|}
%\end{equation}
%$\backslash \backslash$\textbf{Step 5}\;\\
%$dist=\mat{U}\mode {c}*\vec{r}\mode {c}(:,1)$\;\\
%if $dist(1)<dist(2)$ \\
%then print('Unaltered video') \\
%else print('Fake video');
\end{enumerate}
%$\mat{Return:}$ label $i$ that maximizes the above expression
\caption{DeepFake Detection Algorithm}
\label{alg:deepfake}
\end{algorithm}

\begin{figure*}[!bt]
    \begin{center}
  \includegraphics[width = 0.70\textwidth]{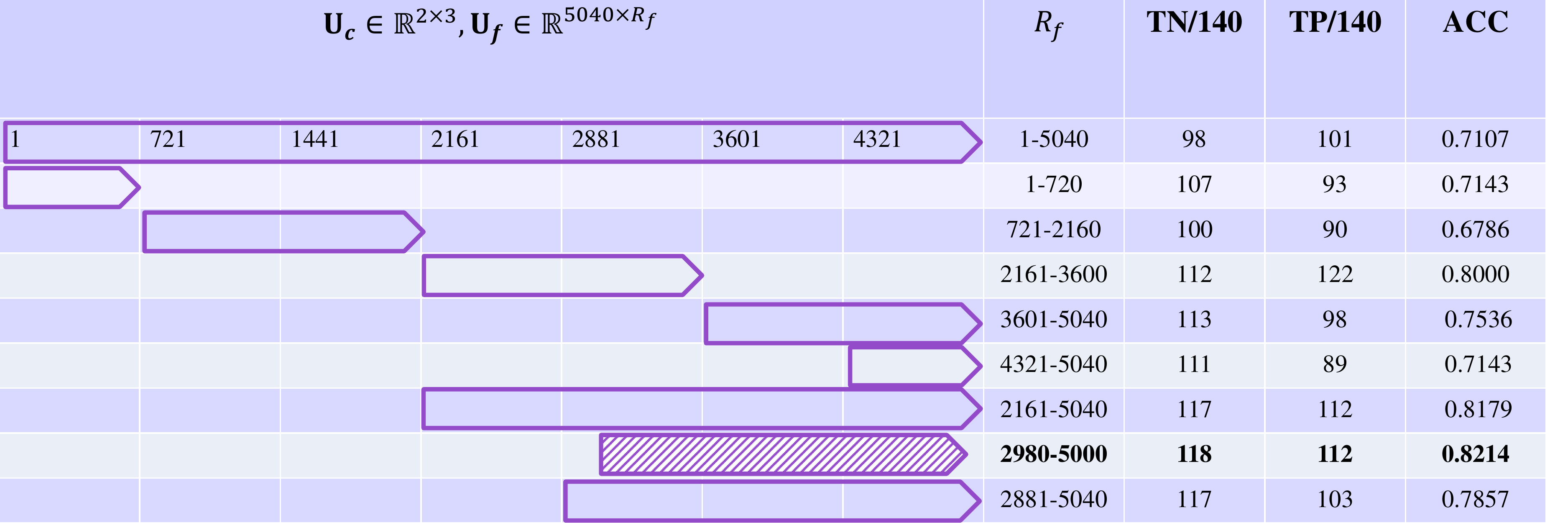}
  %\vspace{-.15in}
    \end{center}
    \caption{Dimensionality reduction experiments video frames compressed with quantization 23. Truncating top $2979$ and bottom $40$ components of the core tensor corresponding to factor matrix $\mat{U}_c$ increases the classification performance. Significant component mostly represent high level structures while insignificant ones may represent noise e.g., artifacts which we want to leverage as a discriminating feature.}
    \label{fig:results}
   % \vspace{-0.2in}
\end{figure*}

\section{Experimental Evaluation}
In this section, we first introduce the dataset and benchmark on this dataset and then we discuss the implementation details and the experimental evaluation.
\subsection{Dataset description}
One of the most popular and widely used databases for image or video forgeries detection is FaceForensics++\footnote{\tiny https://github.com/ondyari/FaceForensics} which first was introduced in 2018 ~\cite{roessler2018faceforensics}. FaceForensics++ comprises more than 500,000 frames from 1000 youtube videos that contain mostly frontal faces~\cite{roessler2019faceforensicspp}. This dataset also includes 1000 videos which are the manipulated version of the original onesand have been  manipulated by four automated face manipulation methods: Deepfakes, Face2Face, FaceSwap and NeuralTextures. 
All original and manipulated videos have constant frame rate of 30 fps and have been compressed lossless with H.264. Moreover, the videos are split up into train set of size 720, validation set of size 140 and test of 140 videos.
binary classification scenario on this dataset. A summarized benchmark of existing techniques on videos manipulated by DeepFakes method is demonstrated in table \ref{table:benchmark}.
The state-of-the-art benchmark on FaceForensics++ is available in  GitHub\footnote{\tiny http://kaldir.vc.in.tum.de/faceforensics-benchmark/}. In this work, we experiment on images manipulated by DeepFake technique.
\begin{table}[t!]
\centering
\small
\begin{tabular}{c c c}
\toprule
\small
\centering
\textbf{Method}&\textbf{Accuracy}\\
\midrule
ZAntiFakeBio&\textbf{1.000}\\	
Leo&\textbf{1.000}\\
Aquarius&\textbf{1.000}\\
RobustForensics&0.991\\	
NoSenseAtAll&0.982\\
PredictFake&0.973\\	
Cancer&0.964\\
Balance&0.918\\
unet+res&0.882\\
HRC&0.827\\
GAEL-Net&0.718\\
\bottomrule
\end{tabular}
\caption{ Summary of Benchmark on FaceForensics++, DeepFakes method~\cite{roessler2019faceforensicspp}}.
\label{table:benchmark}
\vspace{-.2in}
\end{table}

\subsection{Implementation}
 Our work was implemented in MATLAB partially using Tensor Toolbox version 2.6. ~\cite{TTB_Sparse,TTB_Software}. Since all videos have constant frame rate 30 fps, we extracted up to 7 frames for each video by snapping  almost one frame per each 30 seconds using OpenCV library in Python. Moreover, for detecting facial landmarks, we used pretrained dlib face 
 detector\footnote{\tiny http://dlib.net/face\_landmark\_detection.py.html} which is created using the classic Histogram of Oriented Gradients (HOG) feature combined with a linear classifier, an image pyramid, and sliding window detection scheme~\cite{landmark}. For the second step, we calculated the SVD rank r where the r is equal to "$\text{number of train videos}\times7=720\times 7=5040$" for all experiments. 
 The intuition behind this estimation is to have an individual component for each frame.
 Moreover, in contrast to many deep learning approaches for Deepfake detection, our approach does not require GPU base configuration and both train and test steps can be executed on an ordinary CPU based configuration. The description of the CPU based configuration we experimented on is as follows: Intel(R) Core (TM) i5-8600K CPU @3.60GHz,CentOS Linux 7 (Core) operating system and 40GB RAM memory.

\subsection{Evaluation}
\subsubsection{Classification performance}
\par Classification performance of our proposed  multilinear framework when we keep all of the components as well as when we truncate different ranges of components, is illustrated in Figure.~\ref{fig:results}. In this Figure, TN, TP, and ACC. denote true negative, true positive, and accuracy respectively.
\par As demonstrated, truncating top $2980$ and bottom $40$ components, significantly improves the classification accuracy. In this work, we aim to find discriminating representations for outer ring of real vs Deepfake videos introduced by synthesizing artifacts. Thus, we hypothesize the noisy components i.e., components with insignificant values may represent those artifacts. So, by truncating the top components, we avoid high level facial structures and only keep those that correspond to what we aim to capture i.e., artifacts, for the classification. Moreover, the last $40$ components are the most insignificant ones that might be introduced by noises other than synthesizing artifacts. Anyhow, keeping components in range $2980-5000$ results in around $0.82\%$ accuracy.

\subsubsection{Effects of truncation on class representations}
\par
\begin{figure}[!tb]
    \begin{center}
   \subfigure[Before Truncation]{\includegraphics[width=.9\linewidth]{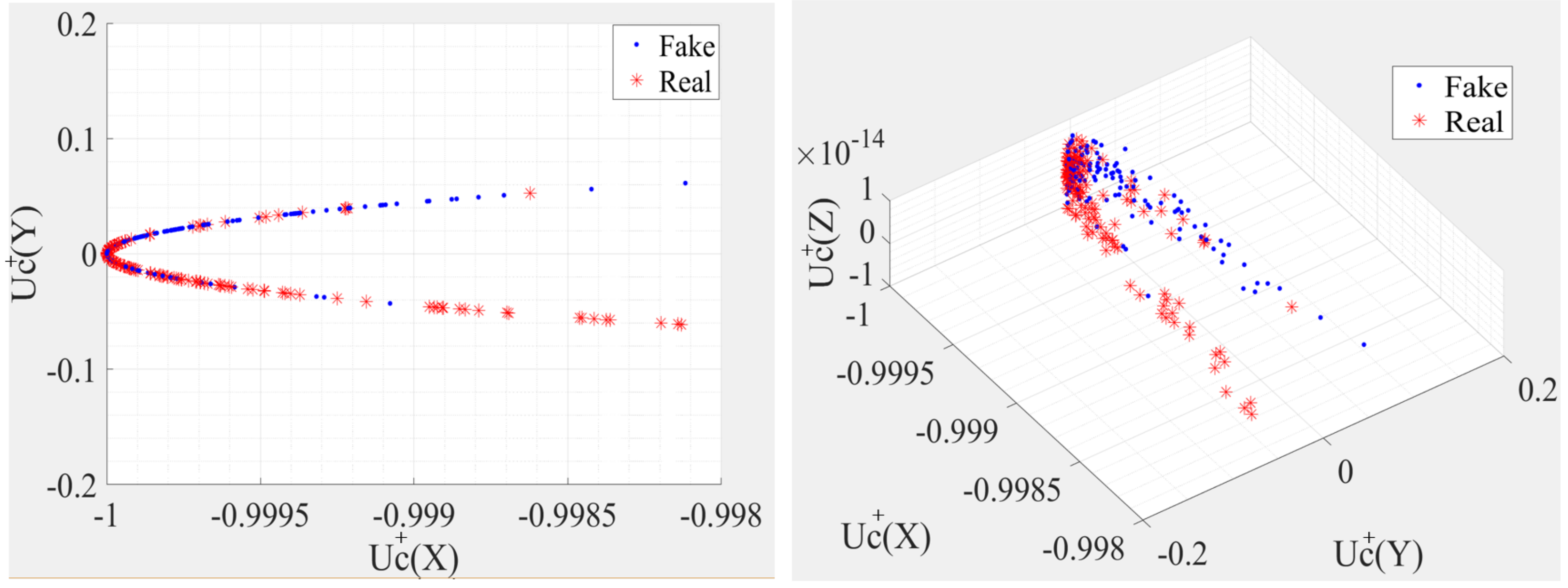}}
\subfigure[After truncation (2980-5000)]{\includegraphics[width = .9\linewidth] {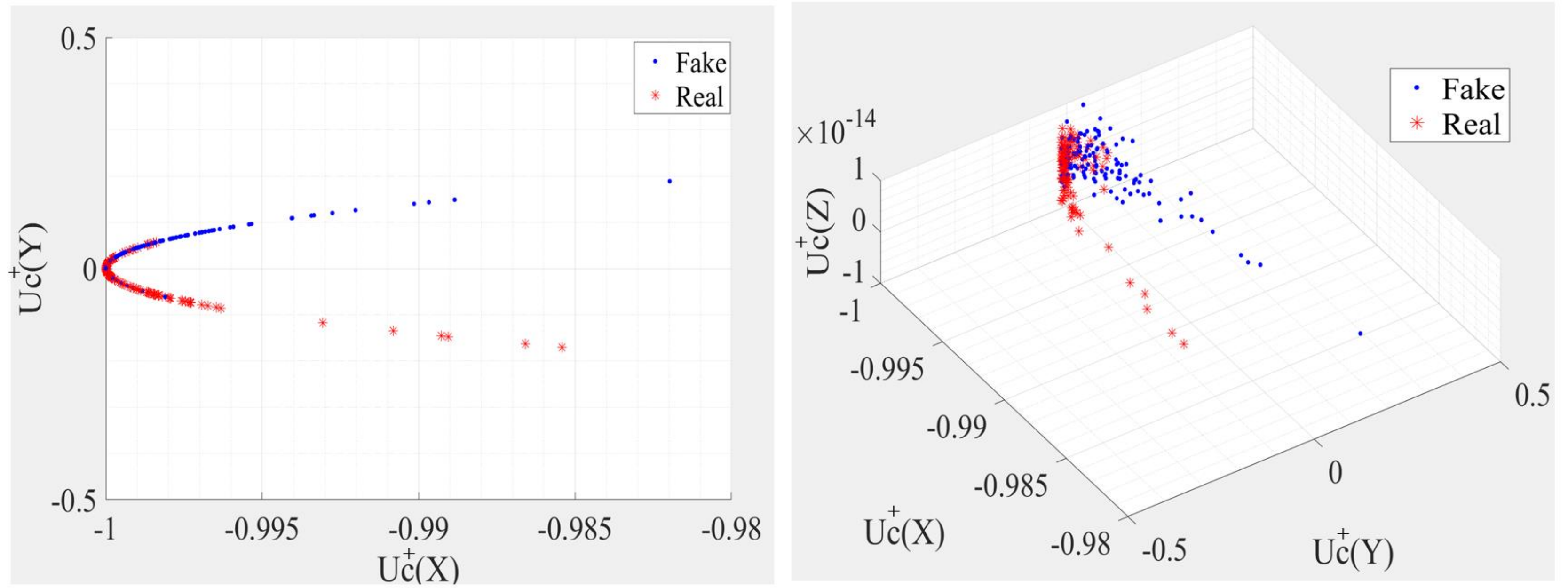}}
    \vspace{-.15in}
    \end{center}
    \caption{Distribution of class coefficients before and after truncation. As illustrated, after truncation, data points of each class get closer to each other and as a result, the number of outliers decreases significantly and and the classes are more linearly separable. In these plots, there are scale differences in the axes but in reality the distributions are nearly straight lines.}
    \label{fig:distribution}
    \vspace{-.15in}
\end{figure}

 To clarify the efficacy of truncation, we depict the PCA coefficients of column vectors of $\mat{U_c^{+}}$ for test frames before and after applying truncation. The distribution of the PCA coefficients is demonstrated in Figure.\ref{fig:distribution}. As shown, truncating the undiscriminating components, makes coefficients of each class more similar and as a result put them closer to each other. Specially in case of samples that are located in outer parts of the semicircle i.e., outliers. In other words, the representations after truncation are more linearly separable than those before applying the truncation. 

\section{ Conclusion and Future Work}
In this work, we leverage the region that we hypothesize has highest concentration of artifacts, the face outer ring, for classification of Deepfakes using our proposed multilinear framework. Our preliminary results show that using only the outer facial ring we achieve 82\% accuracy. 
In future work, we will learn class representations by subdividing an image into parts~\cite{Vasilescu92} and treating them as either 
items 
in a part-based hierarchy %that are analyzed by employing compositional hierarchical tensor factorization~\cite{Vasilescu19},
or
as items 
in a "bag of parts" whose representations may  learned bottom-up~\cite{Vasilescu20}.
%will be performed.
Another direction for future work, is to use binary masks released by~\cite{roessler2019faceforensicspp}. The binary mask can be leveraged for precise segmentation of the frames into regions of interest.

\section{Acknowledgements}
The authors would like to thank Ghazal Mazaheri and Amit Roy-Chowdhury for initial help with the dataset. Research was supported by the National Science Foundation CDS\&E Grant no. OAC-$1808591$ and a UCR Regents Faculty Fellowship. Any opinions, findings, and conclusions or recommendations expressed in this material are those of the author(s) and do not necessarily reflect the views of the funding parties.

%\newpage

\end{document}